%% file: 2015_DeepDiscovery.tex
\documentclass[journal]{IEEEtran}

\usepackage{cite}

\ifCLASSINFOpdf
  \usepackage[pdftex]{graphicx}
\else
\fi
\usepackage{graphicx}
\usepackage{subcaption}
\usepackage{epsfig}
\usepackage{array}
\usepackage{booktabs}
\usepackage{color}
\usepackage{multirow}
\usepackage{url}

\usepackage[cmex10]{amsmath}
\usepackage{amssymb}

\ifCLASSOPTIONcompsoc
  \usepackage[caption=false,font=normalsize,labelfont=sf,textfont=sf]{subfig}
\else
  \usepackage[caption=false,font=footnotesize]{subfig}
\fi

\begin{document}
\title{Discovery Radiomics for Multi-Parametric MRI Prostate Cancer Detection}
\author{Audrey~G.~Chung, \emph{Student~Member, IEEE}, Mohammad~Javad~Shafiee, \emph{Student~Member, IEEE}, Devinder~Kumar, Farzad~Khalvati, Masoom~A.~Haider, and Alexander~Wong, \emph{Member, IEEE}
\thanks{A. G. Chung, M. J. Shafiee, D. Kumar, and A. Wong are are part of
the Vision and Image Processing Research Group with the Department of Systems
Design Engineering, University of Waterloo, Ontario, Canada, N2L 3G1. a28wong@uwaterloo.ca

F. Khalvati and M. A. Haider are with the Department of Medical
Imaging, University of Toronto and Sunnybrook Health Sciences
Centre, Toronto, Ontario, Canada, M4N 3M5. farzad.khalvati@sri.utoronto.ca}}

\markboth{}%
{}

\maketitle

\begin{abstract}
\input{Abstract.tex}
\end{abstract}


\IEEEpeerreviewmaketitle

\section{Introduction}
\label{Introduction}
\input{Introduction.tex}

\section{Methods}
\label{Methods}
\input{Methods.tex}

\section{Results}
\label{Results}
\input{Results.tex}
\section{Conclusion}
\label{Conclusion}
\input{Conclusion.tex}


%

%
%

\section*{Acknowledgements}
This research has been supported by the Ontario Institute of Cancer Research (OICR), Canada Research Chairs programs, Natural Sciences and Engineering Research Council of Canada (NSERC), and the Ministry of Research and Innovation of Ontario. The authors also thank Nvidia for the GPU hardware used in this study through the Nvidia Hardware Grant Program.  %

\section*{Author Contributions}
AC, MJS, and AW contributed to the design and implementation of the concept. AC, MJS, FK, and AW contributed to the design and implementation of the experiments, and performing statistical analysis. DK was involved in performing data augmentation. FK and MH were involved in collecting and reviewing the data. All authors contributed to the writing and reviewing of the paper.

\ifCLASSOPTIONcaptionsoff
  \newpage
\fi



\bibliographystyle{IEEEtran}
%
%
%

\bibliography{DeepDiscovery}

%

%






\end{document}

%% file: Abstract.tex
Prostate cancer is the most diagnosed form of cancer in Canadian men, and is the third leading cause of cancer death. Despite these statistics, prognosis is relatively good with a sufficiently early diagnosis, making fast and reliable prostate cancer detection crucial. As imaging-based prostate cancer screening, such as magnetic resonance imaging (MRI), requires an experienced medical professional to extensively review the data and perform a diagnosis, radiomics-driven methods help streamline the process and has the potential to significantly improve diagnostic accuracy and efficiency, and thus improving patient survival rates. These radiomics-driven methods currently rely on hand-crafted sets of quantitative imaging-based features, which are selected manually and can limit their ability to fully characterize unique prostate cancer tumour phenotype. In this study, we propose a novel \textit{discovery radiomics} framework for generating custom radiomic sequences tailored for prostate cancer detection. Discovery radiomics aims to uncover abstract imaging-based features that capture highly unique tumour traits and characteristics beyond what can be captured using predefined feature models. In this paper, we discover new custom radiomic sequencers for generating new prostate radiomic sequences using multi-parametric MRI data. We evaluated the performance of the discovered radiomic sequencer against a state-of-the-art hand-crafted radiomic sequencer for computer-aided prostate cancer detection with a feedforward neural network using real clinical prostate multi-parametric MRI data. Results for the discovered radiomic sequencer demonstrate good performance in prostate cancer detection and clinical decision support relative to the hand-crafted radiomic sequencer. The use of discovery radiomics shows potential for more efficient and reliable automatic prostate cancer detection.

%

%% file: Introduction.tex
Prostate cancer is the most diagnosed form of cancer (excluding non-melanoma skin cancers) in Canadian and American men. According to the Canadian Cancer Society \cite{CCS2015}, there is an estimated 24,000 new cases and 4,100 deaths from it in 2015, making it the third most deadly cancer and accounting for approximately $10\%$ of cancer deaths in Canadian men. Similarly in the United States, there is an estimated 220,800 new cases and 27,540 deaths from prostate cancer in 2015, making it the second most deadly cancer and accouting for approximately $9\%$ of cancer deaths in American men \cite{ACS2015}. The median patient survival time for metastatic prostate cancer is between 12.2 to 21.7 months \cite{Jemal2011}. However, prognosis is relatively good if the prostate cancer is detected early. As such, fast and reliable prostate cancer screening methods are crucial and can greatly impact patient survival rate, as the five-year survival rate in Canada is $96\%$ for patients diagnosed with prostate cancer before the metastatic stage \cite{CCS2011}.

The current clinical model for initial prostate cancer screening employs a digital rectal exam (DRE) or a prostate-specific antigen (PSA) test. Given a positive DRE or an elevated PSA, a patient undergoes a follow-up transrectal ultrasound (TRUS) guided multicore biopsy for risk stratification. The PSA in particular has recently come under scrutiny, as recent studies \cite{Andriole2009} \cite{Schroder2009} have demonstrated that the PSA test has a significant risk of overdiagnosis with an estimated $50\%$ of screened men being diagnosed with prostate cancer. This oversensitivity leads to expensive and painful needle biopsies and subsequent overtreatment \cite{Andriole2009} \cite{Schroder2009} \cite{Vickers2014}. In addition, these prostate biopsies cause discomfort, possible sexual dysfunction, and increased hospital admission rates due to infectious complications while having a chance of the biopsy needle missing the cancerous tissue \cite{Nam2010} \cite{Loeb2013}. The challenge diagnosticians currently face is how to improve the detection of prostate cancer by reducing the overdiagnosis due to conventional screening methods while still maintaining a high sensitivity.

The use of magnetic resonance imaging (MRI) has recently grown in popularity as a non-invasive imaging-based prostate cancer detection method; however, a diagnosis through MRI requires an experienced medical professional to extensively review the data. Manual labelling of image data is time-consuming, and can lead to diagnostic inconsistencies due to variability between radiologists (inter-observer variability) and the variability of a radiologist over multiple sittings (intra-observer variability) \cite{Jameson2010} \cite{Smith2007} \cite{Asman2011}. To help raise the consistency of radiologists, the European Society of Urogenital Radiology (ESUR) introduced the Prostate Imaging - Reporting And Diagnosis System (PI-RADS) as a common set of criteria \cite{Barentsz2012}. In addition, the use of multiple MRI modalities (multi-parametric MRI) has been shown to improve prostate cancer localization \cite{Haider2007} via the extraction of unique information and features from each modality. Despite PI-RADS and further development to standardize diagnostic practices across multi-parametric MRI \cite{Rothke2013}, there is still a level of subjectiveness in assessing MR images that can lead to inter-observer and intra-observer variability.


\input{Background.tex}

\begin{figure*}[t]
	\centering
	\includegraphics[width=0.85\textwidth]{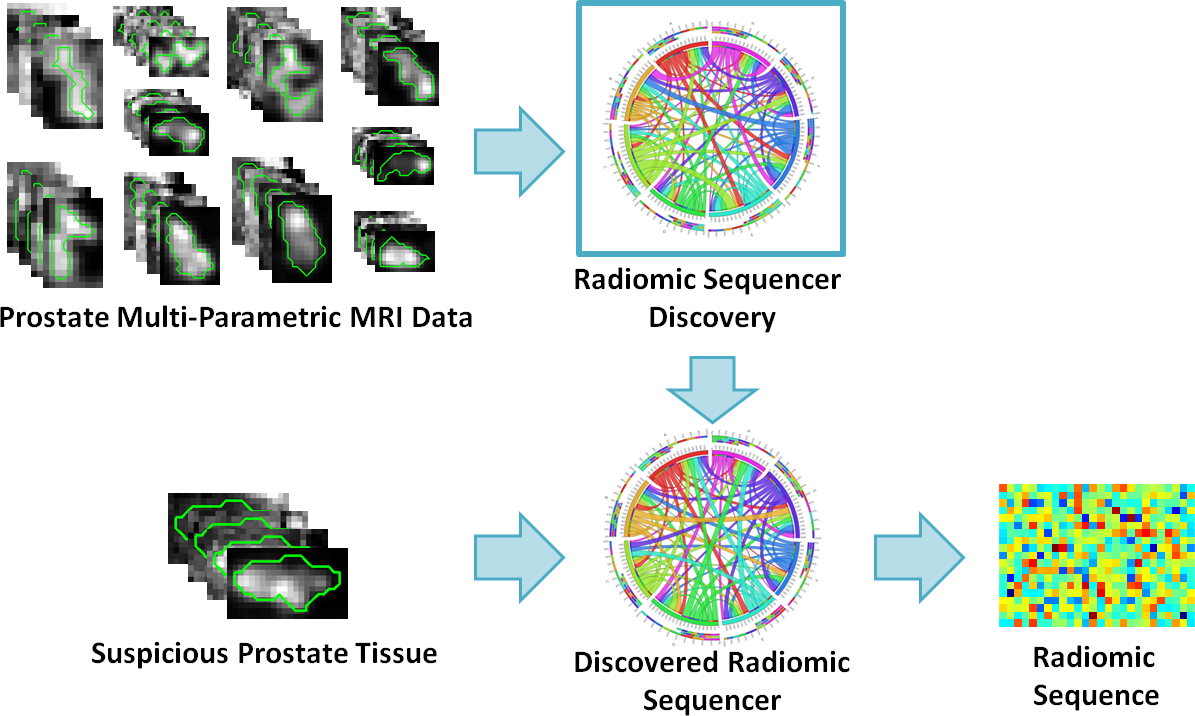}
	\caption{Overview of the proposed discovery radiomics framework for prostate cancer detection.  Multi-parametric MRI data from past patients is used in the radiomic sequencer discovery process to discover a customized radiomic sequencer tailored for prostate cancer. For a new patient case, the patient's multi-parametric MRI data is fed into the discovered radiomic sequencer to generate a custom, tailor-made radiomic sequence of abstract imaging-based features for tumor quantification and analysis.}
	\label{fig_AlgFrame}
\end{figure*}

While hand-crafted imaging-based feature models for radiomics-driven methods for prostate cancer detection have been shown to be highly effective, the generic nature of such feature models for generating radiomic sequences can limit their ability to fully characterize unique prostate cancer tumour phenotype. Motivated by this, we introduce the concept of \textit{discovery radiomics}, where we forgo the notion of predefined feature models by discovering customized, tailored radiomic feature models directly from the wealth of medical imaging data already available.  Discovery radiomics has the potential to find previously unexplored, abstract imaging-based features that capture highly unique tumour traits and characteristics beyond what can be captured using predefined feature models, thus allowing for improved personalized medicine and risk stratification through radiomic sequencing.

To realize the concept of discovery radiomics, we propose a novel framework for discovering customized radiomic sequencers that can generate radiomic sequences of abstract imaging-based features that are tailored for prostate cancer detection using multi-parametric MRI. In this study, we introduce very deep stochastic convolutional radiomic sequencers that are discovered using multi-parametric MRI data captured from past patient cases for generating custom radiomic sequences. The paper is structured as follows. Patient data and the underlying principles of the proposed discovery radiomics framework for prostate cancer detection are described in Section~\ref{Methods}. Experimental setup and comparative results are shown in Section~\ref{Results}. Lastly, conclusions are drawn and future work is discussed in Section~\ref{Conclusion}.

%% file: Background.tex
Automatic computer-aided prostate cancer detection or \textit{radiomics}-driven methods for prostate cancer detection have been developed to help streamline the diagnostic process and increase diagnostic consistency. Radiomics refers to the high-throughput extraction and analysis of large amounts of quantitative features from medical imaging data to characterize tumour phenotypes, allowing for a high-dimensional mineable feature space that can be utilized for cancer detection and prognosis \cite{Lambin2012}. The prognostic power of radiomics has been previously shown in studies on lung and head-and-neck cancer patients, indicating the potential of radiomic features for personalized medicine and predicting patient outcomes \cite{Gevaert2012} \cite{Aerts2014}.

Current radiomic-driven methods for prostate cancer detection typically employ a set of pre-defined, hand-crafted quantitative features extracted from multi-parametric MR images. Lema{\^\i}tre \textit{et al.} \cite{Lemaitre2015} recently published a comprehensive review of state-of-art radiomics-driven methods for prostate cancer detection and diagnosis. The hand-crafted feature models used in these radiomics-driven methods typically included some combination of the following: first-order and second-order statistical features, Gabor filters, gradient-based features, fractal-based features, pharmacokinetic features, and discrete cosine transform (DCT) features.

Madabhushi \textit{et al.} \cite{Madabhushi2005} assessed the utility of combining multiple features for detecting prostate cancer in \textit{ex-vivo} MRI (i.e., prostates glands obtained via radical prostatectomy). Three-dimensional texture feature sets were extracted from MRIs that had been corrected for background inhomogeneity and nonstandardness. The following feature sets were extracted and used to train an ensemble of classifiers: first-order statistical features, second-order Haralick features, steerable Gabor filters, and gradient-based features.

Tiwari \textit{et al.} \cite{Tiwari2013} proposed a method that combines structural and metabolic imaging data for prostate cancer detection in multi-parametric MRI (including T2-weighting imaging and magnetic resonance spectroscopy). Using similar features as \cite{Madabhushi2005}, Tiwari \textit{et al.} detected cancerous regions within prostate tissue using a random forest classifier.

Duda \textit{et al.} \cite{Duda2014} introduced a semi-automatic multi-image texture analysis for the characterization of prostate tissue using contrast-enhanced T1-weighted, T2-weighted, and diffusion-weighted imaging. The method simultaneously analysed several images (each acquired under different conditions) representing the same part of the organ. In addition to the features used by \cite{Madabhushi2005}, Duda \textit{et al.} also extracted fractal-based and run length features.

Litjens \textit{et al.} \cite{Litjens2014} motivated their features with biology, using features that represent pharmacokinetic behaviour, symmetry and appearance, and other anatomical aspects. Various features were extracted from different MR images: second-order statistical and Gabor features from T2-weighted images, multi-scale blobness from apparent diffusion coefficient maps, and curve fitting and pharmacokinetic features from contrast enhanced images. These features were used for prostate gland segmentation, generating a cancer likelihood map, and cancerous region classification.

Ozer \textit{et al.} \cite{Ozer2009} also proposed the use of pharmacokinetic parameters derived from contrast-enhanced MRI, combining it with T2-weighted and diffusion-weighted imaging. Using a relevance vector machine (RVM) with a Bayesian framework, Ozer \textit{et al.} leveraged second-order statistical and (DCT) features from the peripheral zone of multi-parametric prostate MRI datasets to automatically segmented regions of cancerous tissue, and evaluated the method against support vector machines (SVM) with the same framework. Ozer \textit{et al.} later extended their work to select a threshold value for increased segmentation performance, and further compared with a representative unsupervised segmentation method (Markov random field) \cite{Ozer2010}.

Artan \textit{et al.} \cite{Artan2010} hand-crafted feature vectors using median-filtered intensity values extracted from axial-oblique fast spin-echo (FSE) T2-weighted, echo planar diffusion-weighted, multi-echo FSE, and contrast-enhance MR images. Using these features, Artan \textit{et al.} developed a cost-sensitive SVM for automated prostate cancer localization and showed improved cancer localization accuracy over conventional SVMs. They also combined a conditional random field with the cost-sensitive framework that further improved prostate cancer localization via the incorporation of spatial information.

Similar to \cite{Ozer2010}, Liu \textit{et al.} \cite{Liu2009} introduced a method for unsupervised prostate cancer segmentation using fuzzy Markov random fields. They estimated the parameters of the Markovian distribution of the measured data, and applied it to parameter maps extracted from multi-parametric prostate MRI (T2-weighted MRI, quantitative T2, diffusion-weighted imaging, and contrast-enhanced MRI).

Vos \textit{et al.} \cite{Vos2012} developed a fully automatic computer-aided detection method for prostate cancer using a supervised classifier in a two-stage classification approach. As prostate cancer can be discriminated from benign abnormalities due to their heterogeneity, Vos \textit{et al.} analysed lesion candidates via the combination of a histogram analysis of T2-weighted axial images, pharmacokinetic maps, contrast-enhanced T1-weighted, and apparent diffusion coefficient maps with texture-based features.

Khalvati \textit{et al.} \cite{Khalvati2014} proposed a multi-parametric MRI texture feature model for radiomics-driven prostate cancer analysis. The texture feature model, based on the one proposed by Peng \textit{et al.} \cite{Peng2013}, comprises of 19 low-level texture features extracted from each MRI modality, including features extracted from the gray-level co-occurrence matrix (GLCM). Khalvati \textit{et al.} \cite{Khalvati2015} more recently published radiomics-driven models as an extension of the previous texture feature model. An attempt at designing comprehensive quantitative feature sequences, the radiomics-driven models include additional MRI modalities, additional low-level features, and feature selection.

%% file: Methods.tex
\begin{figure*}[t]
	\centering
	\includegraphics[width=\textwidth]{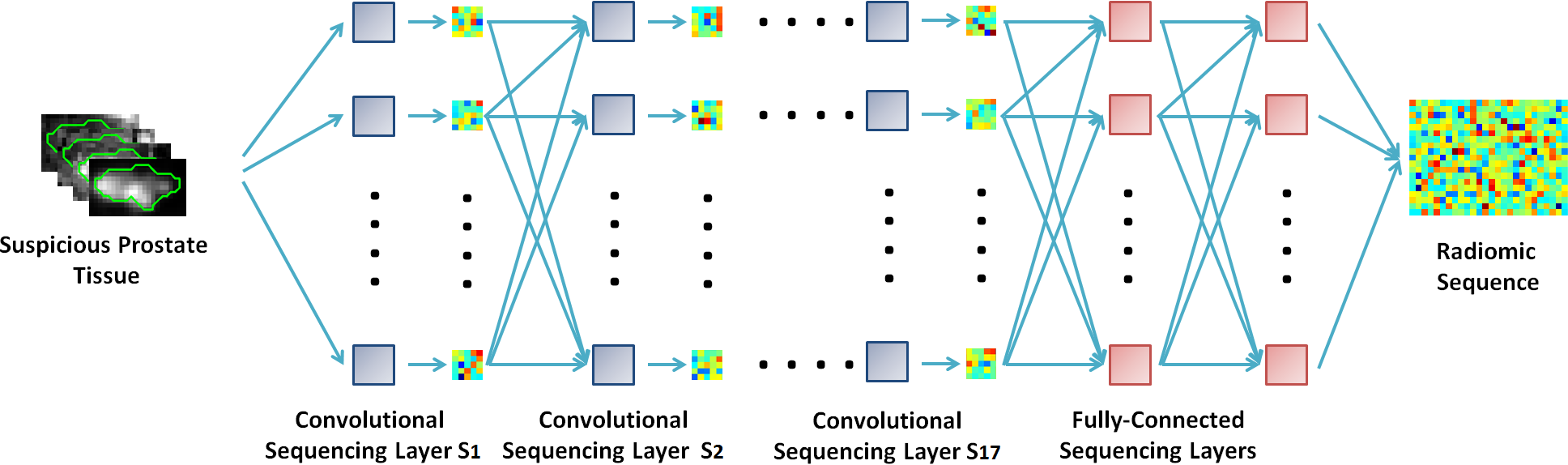}
	\caption{Architecture of the very deep stochastic convolutional radiomic sequencer introduced in this study.  The radiomic sequencer consists of 17 convolutional sequencing layers and two fully-connected sequencing layers.  The weights of the connections are determined stochastically based on a learned distribution.}
	\label{fig_ConvFramework}
\end{figure*}

The proposed discovery radiomics framework for prostate cancer detection consists of the following steps (see Figure~\ref{fig_AlgFrame}). First, standardized multi-parametric MRI data from past patients along with pathology-verified radiologist tissue annotations are fed into the radiomic sequencer discovery process, where a customized radiomic sequencer is discovered for generating a large number of abstract imaging-based features that capture highly unique tumour traits and characteristics pertaining to prostate cancer. Second, for a new patient case, the discovered radiomic sequencer is then used to generate a custom, tailor-made radiomic sequence from the multi-parametric MRI data of the new patient case for comprehensive, custom quantification of the prostate cancer tumour phenotype. These discovered radiomic sequences can then be used by a classifier to provide information for clinical decision support for prostate cancer detection and diagnosis.

\subsection{Patient Data Collection}
A requirement of the proposed discovery radiomics framework is the presence of standardized multi-parametric MRI data from past patients. In this study, multi-parametric MRI data of 20 patients was acquired using a Philips Achieva 3.0T machine at Sunnybrook Health Sciences Centre, Toronto, Ontario, Canada. Institutional research ethics board approval and patient informed consent for this study was obtained by the Research Ethics Board of Sunnybrook Health Sciences Centre. The patients' ages ranged from 53 to 83. T2-weighted (T2-w), diffusion-weighted imaging (DWI), and correlated diffusion imaging (CDI) axial data was collected for each patient to form the multi-parametric MRI dataset. Table~\ref{tab_Data} summarizes the information regarding the acquired prostate multi-parametric MRI data including displayed field of view (DFOV), resolution, echo time (TE), and repetition time (TR).

\begin{table}[h]
	\centering
	\caption{Description of acquired prostate T2w, DWI, and CDI imaging data comprising the multi-parametric MRI dataset.}
	\begin{tabular}{c|c|c|c|c}
		\hline
		\textbf{Modality} & \textbf{DFOV ($cm^2$)} & \textbf{Resolution ($mm^3$} &	\textbf{TE ($ms$)} &	\textbf{TR ($ms$)} \\
		\hline
		T2w	& 	$22 \times 22$  &	$0.49 \times 0.49 \times 3$ 	&	$110$	&		$4,697$		\\
		DWI	& 	$20 \times 20$  &	$1.56 \times 1.56 \times 3$ 	&	$61$	&		$6,178$		\\
		CDI	& 	$20 \times 20$  &	$1.56 \times 1.56 \times 3$ 	&	$61$	&		$6,178$		\\
		\hline
	\end{tabular}
	\label{tab_Data}
\end{table}	

The multi-parametric MRI dataset also includes annotation information for isolating the prostate gland, and ground truth data for the size and location of tumour candidates. As ground truth, all images were reviewed and marked as healthy and cancerous tissue by a radiologist with 18 and 13 years of experience interpreting body and prostate MRI, respectively. The multi-parametric MRI data and radiologist annotations were verified as accurate using corresponding histopathology data obtained through radical prostatectomy with Gleason scores of seven or higher. The study used the following multi-parametric MRI modalities: T2-weighted (T2w) imaging, apparent diffusion coefficient (ADC) maps, computed high-b diffusion-weighted imaging (CHB-DWI), and correlated diffusion imaging (CDI). The modalities are summarized in the following subsections.

\subsubsection{T2-weighted Imaging (T2w)}
T2-weighted (T2w) imaging is a MR imaging modality that characterizes the sensitivity of tissue using the differences in transverse (spin-spin) relaxation time of the applied magnetic field. T2w imaging has been shown to provide some localization information due to a small reduction in signal for cancerous tissue in the prostate gland \cite{Haider2007}.

\subsubsection{Diffusion-Weighted Imaging (DWI)}
Diffusion-weighted imaging (DWI) is an imaging modality in which the sensitivity of the tissue to the Brownian motion water molecules is measured through the application of lobe gradients (pairs of opposing magnetic field gradient pulses) \cite{Koh2014}. The diffusion-weighted signal $S$ is formulated as:
\begin{align}
	S = S_0e^{-bD}
\end{align}
\noindent where $S_0$ is the signal intensity without diffusion weighting, $b$ is the gradient strength and pulse duration, and $D$ is the strength of the diffusion. The diffusion-weighted images are typically generated using different $b$ values, and can be used to estimate apparent diffusion coefficient (ADC) maps via least-squares or maximum likelihood methods \cite{Koh2014}. Cancerous tissue in ADC maps is usually presented with a darker intensity relative to surrounding healthy tissue \cite{Walker-Samuel2009}.

\subsubsection{Computed High-b Diffusion Weighted Imaging (CHB-DWI)}
Previous research has shown that high b-values in DWI data (e.g., b-values greater than $1,000 s/mm^2$) allow for increased delineation between healthy and cancerous tissue \cite{Glaister2012} \cite{Rosenkrantz2013}. Due to hardware limitations, acquiring high b-value images for prostate imaging is infeasible. CHB-DWI is a computational model for reconstructing high b-value DWI data using low b-value acquisitions \cite{Glaister2012} \cite{Shafiee2015}. Our patient data includes CHB-DWI images constructed at a b-value of $2,000 s/mm^2$ using a Bayesian model with the least-squares estimation used to estimate our ADC maps.

\subsubsection{Correlated Diffusion Imaging (CDI)}
Correlated Diffusion Imaging (CDI) is a new diffusion MRI modality that leverages the joint correlation in signal attenuation across multiple gradient pulse strengths and timings to improve delineation between cancerous and healthy tissue \cite{Wong2013}. As cancerous tissue generates higher intensities at high b-values, better delineation can be achieved by adjusting the utilized b-values for a given application. The overall characterization of the water diffusion is better represented via the correlation of signal attenuation across all b-values within a local sub-volume, which is obtained via signal mixing \cite{Wong2013}:
\begin{align}	
	CDI(x) = \int ... \int_{b0}^{bn} S_0(x) ... S_n(x)P(S_0(x), ... S_n(x) | \nonumber \\
			 V(x) \times dS_0(x) ... dS_n(x)
\end{align}
\noindent where $b_i$ represents the utilized b-values, $x$ is the spatial location, $S$ is the acquired signals, $P$ represents the conditional joint probability density function, and $V(x)$ is the local sub-volume centred at $x$.

\subsection{Tumour Candidate Detection}
As CDI \cite{Wong2013} shows excellent delineation between cancerous and healthy prostate tissue, tumour candidates were extracted from the dataset using a simple threshold on CDI data:
\begin{align}
	CDI(x)_{mask} =
	\begin{cases}
	1 & CDI(x) > \frac{CDI_{max}}{2} \\
	0 & otherwise
	\end{cases}
\end{align}
\noindent where $CDI(x)$ refers to the CDI value at $x$, $CDI_{max}$ is the max CDI value, and $CDI_{mask}$ is the binary mask of tumour candidates for a given MRI slice. A total of 80 cancerous regions and 714 healthy regions were identified as tumour candidates from the 20 different patients and used in this study.

\subsection{Radiomic Sequencer Discovery}
Given the standardized multi-parametric MRI data from past patients and pathology-verified radiologist tissue annotations, the radiomic sequencer discovery process discovers a customized radiomic sequencer for generating a large number of abstract imaging-based features that capture highly unique tumour traits and characteristics pertaining to prostate cancer. In this study, we introduce a novel very deep stochastic convolutional radiomic sequencer, as shown in Figure~\ref{fig_ConvFramework}, for discovering custom-tailored radiomic sequences for prostate cancer detection and diagnosis. Inspired by the very deep convolutional network structure in \cite{Simonyan2014}, the proposed radiomic sequencer consists of 17 convolutional sequencing layers and two fully-connected sequencing layers of 1000 and 500 nodes, respectively. An important challenge with conventional very deep convolutional networks is that they have a large number of parameters to learn, necessitating an extremely large dataset for training. Due to the limited quantity of prostate cancer patient cases that can be obtained, such networks cannot be learned sufficiently and thus, it is difficult to use such networks as radiomic sequencers. We address this important challenge in two main ways in the proposed radiomic sequencer.

First and foremost, we forgo the notion of having the weights of the connections between nodes be individual parameters to learn, and instead have the weights of the connections between nodes determined stochastically based on a learned distribution. As such, the number of parameters that must be learned is significantly reduced, making it well-suited as a radiomic sequencer that can be discovered with a reasonable amount of patient cases. Second, data augmentation is performed via the rotation of each tumour candidate at $45^\circ$ intervals (i.e., $0^\circ$, $45^\circ$, $90^\circ$, $135^\circ$, $180^\circ$, $225^\circ$, $270^\circ$, $315^\circ$). While the proposed radiomic sequencer has significantly fewer parameters to learn relative to conventional networks, the current dataset is still insufficiently large to properly train the radiomic sequencer. Therefore, data augmentation is used to grow our existing dataset, resulting in 640 cancerous regions and 5,712 healthy regions (as determined by corresponding radiologist annotations and pathology data) that can now be used as tumour candidates for training the proposed radiomic sequencer. In this study, the proposed radiomic sequencer is discovered via iterative scaled conjugate gradient optimization using cross-entropy as the objective function.

\begin{figure}[h]
	\centering
	\includegraphics[width=\linewidth]{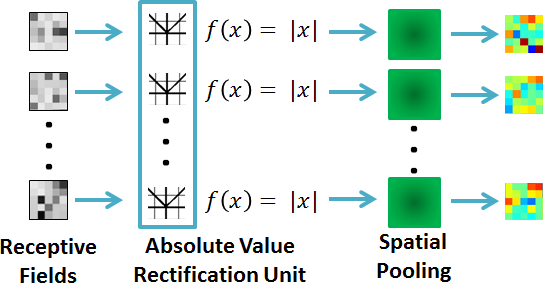}
	\caption{Each convolutional sequencing layer consists of stochastically realized receptive fields, absolute value rectification unit (AVReU), and spatial overlapping median pooling.}
	\label{fig_SeqUnit}
\end{figure}

In the proposed radiomic sequencer, each convolutional sequencing layer (Figure~\ref{fig_SeqUnit}) consists of stochastically realized receptive fields, an absolute value rectification unit (AVReU) to introduce non-saturating nonlinearity into the sequencer, and a spatial overlapping median pooling layer. Detailed descriptions of each part of the convolutional sequencing layer is described in the following subsections.

\subsubsection{Stochastically Realized Receptive Fields}
Unlike conventional convolutional networks where the weights of receptive field are trained via back propagation \cite{Krizhevsky2012}, the proposed radiomic sequencer leverages stochastically realized receptive fields to significantly reduce the number of parameters that need to be learned.  The connection weights encapsulated by the receptive fields are abstracted into probability distributions, and each receptive field represents the shared weights for all localized nodal collections.  In this study, each receptive field is treated as a random field, with the connection weights in the receptive field being spatially correlated subject to a spatial correlation parameter $\psi$, which varies the level of spatial correlation between connection weights within a receptive field.  As such, rather than learning individual connection weights for the receptive fields, one now only needs to learn a single parameter $\psi$.  The proposed radiomic sequencer has 17 convolutional sequencing layers; the number of stochastically realized receptive fields and receptive field size used in each convolutional sequencing layer is specified in Table~\ref{tab_Kernels}.

\begin{table}[h]
	\centering
	\caption{Summary of number of stochastically realized receptive fields and receptive field size at each convolutional sequencing layer.}
	\begin{tabular}{>{\centering\arraybackslash}m{2.5cm}|>{\centering\arraybackslash}m{2.5cm}|>{\centering\arraybackslash}m{2.5cm}}
		\hline
		\textbf{Convolutional Sequencing Layer} & \textbf{Number of Receptive Fields} & \textbf{Receptive Field Size} \\
		\hline
		1, 2							& 	$64$  		&	$5 \times 5$	\\
		3, 4							& 	$128$  		&	$5 \times 5$	\\
		5, 6, 7, 8						& 	$256$  		&	$5 \times 5$	\\
		9, 10, 11, 12, 13, 14, 15, 16	& 	$512$  		&	$5 \times 5$	\\
		17								& 	$2000$  	&	$5 \times 5$	\\
		\hline
	\end{tabular}
	\label{tab_Kernels}
\end{table}	

\subsubsection{Absolute Value Rectification Unit}
Conventional convolutional networks typically model a nodes's output $f$ as a function of its input $x$, such that $f(x) = tanh(x)$ or $f(x) = (1 + e^{-x})^{-1}$. However, these saturating nonlinearities train considerably slower than non-saturating nonlinearities (generally referred to as Rectified Linear Units or ReLUs) such as $f(x) = max(0,x)$. Similar to Krizhevsky \textit{et al.} \cite{Krizhevsky2012}, we incorporate absolute value rectification units (AVReUs) ($f(x) = |x|$) to introduce non-saturating nonlinearity into the proposed radiomic sequencer.

\subsubsection{Spatial Pooling}
Traditional local pooling layers in convolutional neural networks encapsulate the outputs of neighbouring nodes, and are typically done by pooling non-overlapping neighbourhoods into a single value centred at the location of the pooling neighbourhood \cite{Krizhevsky2012}. Non-overlapping spatial pooling is usually performed to reduce computational complexity by indirectly decreasing the effective image size after each pooling operation.  Since with the proposed radiomic sequencer the parameters are already significantly reduced, we instead introduce a spatial non-overlapping median pooling layer in each convolutional sequencing layer that preserves the effective image size.

%% file: Results.tex
\subsection{Experimental Setup}
To assess the usefulness of the proposed discovery radiomics framework for computer-aided prostate cancer detection, the discovered radiomic sequencer (DRS) from the proposed framework was compared against Peng \textit{et al.}'s \cite{Peng2013} and Khalvati \textit{et al.}'s \cite{Khalvati2015} hand-crafted radiomic sequencers for classifying tumour candidates as either healthy or cancerous using a feedforward neural network classifier with a single hidden layer of 100 nodes. While \cite{Peng2013} and \cite{Khalvati2015} evaluated the hand-crafted radiomic sequencers for voxel-based classification via a linear discriminant analysis (LDA) classifier and support vector machine (SVM), respectively, only the hand-crafted radiomic sequencers themselves were included in this study to assess the use of hand-crafted and discovered radiomic sequencers. The performance of the classifier learned based on each radiomic sequencer was evaluated using leave-one-patient-out cross-validation using balanced training and testing data (i.e., an equal number of healthy and cancerous tumour candidates), and quantitatively assessed via sensitivity, specificity, and accuracy metrics:

\begin{align*}
	\textrm{Sensitivity} = \frac{TP}{P}\hspace{10pt}&\hspace{10pt}
	\textrm{Specificity} = \frac{TN }{N}\\ \\
	\textrm{Accuracy}&=\frac{TN+TP}{N+P}
\end{align*}
\vspace{0.1mm}

\noindent where the performance of each method was quantified by the metrics' closeness to one. TP is the number of tumour candidates identified as cancerous by both the classifier and the radiologist's tissue segmentation, TN is the number of tumour candidates identified as healthy by both the classifier and the radiologist's segmentation, N is the number of tumour candidates not in the radiologist segmented tissue (i.e., healthy prostate tissue), and P is the number of tumour candidates in the radiologist segmented tissue (i.e., cancerous tissue).

To further analyse the effectiveness of DRS relative to Peng \textit{et al.} \cite{Peng2013} and Khalvati \textit{et al.} \cite{Khalvati2015}, the Fisher criterion was used to assess the separability of the generated imaging-based features. The Fisher Criterion (FC) represents the separability of the generated healthy and cancerous radiomic sequences. The FC scores for the radiomic sequences generated for healthy (h) and cancerous (c) candidate regions were computed via the following:

\begin{align}
	FC = \frac{(\mu_h - \mu_c)^2}{\sigma_h + \sigma_c}
\end{align}

\noindent where $\mu_h$ and $\mu_c$ represent the mean values of the healthy and cancerous radiomic sequences, respectively, and $\sigma_h$ and $\sigma_c$ represent their respective standard deviations. A higher FC score corresponds to a higher degree of separability between the healthy and cancerous radiomic sequences, indicating the better delineation between healthy and cancerous regions using the generated radiomic sequences.

\subsection{Results}
The discovered radiomic sequencer (DRS) was evaluated against two state-of-the-art hand-crafted radiomic sequencers proposed by Peng \textit{et al.} \cite{Peng2013} and Khalvati \textit{et al.} \cite{Khalvati2015} for classifying tumour candidates as either healthy or cancerous using a feedforward neural network classifier with a single hidden layer of 100 nodes. Sensitivity, specificity, and accuracy were calculated via leave-one-patient-out cross-validation with balanced training and testing datasets (as the majority of the tumour candidates were healthy) to prevent the accuracy metric from being strongly skewed by the specificity rate. Table~\ref{tab_Results1} shows the performance metrics for classification using the hand-crafted radiomic sequencers \cite{Peng2013} \cite{Khalvati2015} and the discovered radiomic sequencer.

\begin{table}[h]
	\centering
	\caption{Comparison of hand-crafted radiomic sequencers \cite{Peng2013} \cite{Khalvati2015} with discovered radiomic sequencer (DRS) for tumour candidate classification.}
	\begin{tabular}{c|c|c|c}
		\hline
		& \textbf{Sensitivity} & 	\textbf{Specificity} &		\textbf{Accuracy} \\
		\hline
		
		\textbf{Peng \textit{et al.}} \cite{Peng2013}				& 	$0.1824$  		&	$0.9005$ 		&	$0.5821$	\\
		\textbf{Khalvati \textit{et al.}} \cite{Khalvati2015}		& 	$0.3568$  		&	\textbf{0.9231} &	$0.6730$	\\
		\textbf{DRS}								& 	\textbf{0.6400}	&	$0.8203$		&	\textbf{0.7365}		\\ 
		\hline
	\end{tabular}
	\label{tab_Results1}
\end{table}	

As shown in Table~\ref{tab_Results1}, Khalvati \textit{et al.}~\cite{Khalvati2015} performed better than Peng \textit{et al.}~\cite{Peng2013} in terms of all performance metrics, and produced the highest specificity at $92.31\%$. While \cite{Khalvati2015} has a high specificity, it is worth noting that the hand-crafted radiomic sequencer also has the lowest sensitivity (i.e., proportion of correctly identified cancerous tumour candidates) at $35.68\%$, missing almost two thirds of the cancerous candidates. This indicates that the hand-crafted radiomic sequencer generates radiomic sequences that better represent tumour candidates consisting of healthy tissue than tumour candidates consisting of cancerous tissue.

Table~\ref{tab_Results1} also shows that the discovered radiomic sequencer produced the highest sensitivity at $64.00\%$ and accuracy at $73.65\%$. In addition, the discovered radiomic sequencer has noticeably more consistent performance across the metrics relative to both hand-crafted radiomic sequencers \cite{Peng2013} \cite{Khalvati2015}, with specificity at $82.03\%$. This suggests that the custom radiomic sequences generated from the discovered radiomic sequencer are better able to represent both healthy and cancerous prostate tissue in a more balanced fashion, as opposed to favouring healthy tissue.

Statistical significance for each of the performance metrics was performed via a paired two-sided null hypothesis test (as shown in Table~\ref{tab_Results2}). The improvements in sensitivity using the discovered radiomic sequencer is statistically significant compared to both \cite{Peng2013} and \cite{Khalvati2015}, with the improvements in accuracy using the discovered radiomic sequencer statistically significant compared to \cite{Peng2013}. In addition, a separability analysis was performed using the Fisher Criterion; the results of the separability analysis are shown in Table~\ref{tab_Separability}.	

\begin{table}[h]
	\centering
	\caption{Paired two-sided null hypothesis test results for statistical significance between hand-crafted radiomic sequencers \cite{Peng2013} \cite{Khalvati2015} and discovered radiomic sequencer (DRS).}
	\begin{tabular}{c|c|c|c}
		\hline
		& \textbf{Sensitivity} & 	\textbf{Specificity} &		\textbf{Accuracy} \\
		\hline
		
		\textbf{p-value against} \cite{Peng2013}	& 	$0.0029$  		&	$0.0165$ 		&	$0.0133$	\\
		\textbf{p-value against} \cite{Khalvati2015}& 	$0.0277$  		&	$0.0056$ 		&	$0.2396$	\\
		\hline
	\end{tabular}
	\label{tab_Results2}
\end{table}

Table~\ref{tab_Separability} shows that DRS obtained the highest FC score relative to Peng \textit{et al.}~\cite{Peng2013} and Khalvati \textit{et al.}~\cite{Khalvati2015}. This indicates that the discovered radiomic sequences produced by DRS has the best separability between healthy and cancerous radiomic sequences. These results demonstrate the potential of the proposed radiomics discovery framework for building custom radiomic sequencers that can generate radiomic sequences tailored for prostate cancer characterization and detection.

\begin{table}[h]
	\centering
	\caption{Comparison of hand-crafted radiomic sequencers \cite{Peng2013} \cite{Khalvati2015} with discovered radiomic sequencer (DRS) with respect to separability of extracted features.}
	\begin{tabular}{c|c}
		\hline
		& \textbf{Fisher Criterion} \\
		\hline		
		\textbf{Peng \textit{et al.}} \cite{Peng2013}				& 	$0.0188$  		\\
		\textbf{Khalvati \textit{et al.}} \cite{Khalvati2015}		& 	$0.0144$  		\\
		\textbf{DRS}												& 	\textbf{0.0712}	\\
		\hline
	\end{tabular}
	\label{tab_Separability}
\end{table}	

%% file: Conclusion.tex
In this study, a novel discovery radiomics framework for prostate cancer detection using multi-parametric MRI data was presented. Unlike conventional radiomics-driven methods that use a set of hand-crafted radiomic features, the discovered radiomic sequencer can generate radiomic sequences that are specifically tailored for quantifying and differentiating healthy and cancerous prostate tissue.

The performance of the discovered radiomic sequencer was compared against two state-of-the-art hand-crafted radiomic sequencers \cite{Peng2013} \cite{Khalvati2015} via leave-one-patient-out cross-validation for the task of tumour candidate classification using a feedforward neural network. While \cite{Khalvati2015} produced the highest specificity ($92.31\%$) and outperformed \cite{Peng2013} with respect to all metrics, the Khalvati \textit{et al.}'s hand-crafted radiomic sequencer also produced the second lowest sensitivity ($35.68\%$), indicating that almost two thirds of cancerous candidates are undetected. The discovered radiomic sequencer based on the proposed discovery radiomics framework, however, produced the highest sensitivity ($64.00\%$) and accuracy ($73.65\%$), and has noticeably more consistent performance across the metrics relative to \cite{Peng2013} and \cite{Khalvati2015}, with specificity at $82.48\%$.

Future work includes the use of additional MRI modalities (such as dual-stage correlated diffusion imaging \cite{Wong2015}) to allow the proposed framework to discover radiomic sequencers that can produce more distinctive radiomic sequences for prostate cancer, and the investigation of transfer learning \cite{Pan2010} (e.g., the application of a radiomic sequencer trained using non-prostate imaging data to prostate cancer data). In addition, further analysis of discovered sequences will be conducted to better understand the prostate cancer phenotype. The application of radiomic sequences can be extended to cancer grading and staging, enabling a non-invasive method for assessing the severity and behaviour of prostate cancer. As such, radiomic sequencing can lead to better patient care and survival rates through more reliable risk stratification.